\documentclass[10pt,letterpaper]{article}

\usepackage{spatialbench-preprint}

\preprintKicker{bioRxiv preprint}
\preprintCategory{Systems Biology}
\preprintDate{May 2026}
\preprintTitle{Verifiable Benchmarking of Long-Horizon Spatial Biology}
\preprintSubtitle{Evaluating whether AI agents can recover complex scientific conclusions from raw spatial biology data}
\preprintAuthors{Ian Diks\preprintSup{1,*}, Harihara Muralidharan\preprintSup{1,*}, Tim Proctor\preprintSup{1}, Kenny Workman\preprintSup{1,*}}
\preprintAffiliations{\preprintSup{1}LatchBio, San Francisco, CA, USA\\\preprintSup{*}Equal contribution}
\preprintCorrespondence{kenny@latch.bio}

\begin{document}

\PreprintHeader

\begin{PreprintAbstract}

AI agents are increasingly useful for biological data analysis, but existing
benchmarks mostly test broad biological knowledge, executable workflows, or
localized analysis steps rather than end-to-end scientific reasoning over
spatial measurements. We introduce SpatialBench-Long, a benchmark for
long-horizon spatial biology in which agents must recover biological claims
from raw or near-raw data and calibrated experimental context without
prescribed methods. SpatialBench-Long contains 24 evaluations across primary
pancreatic ductal adenocarcinoma (PDAC), engineered glioblastoma organoids and
in vivo tumors, Cas9 lineage-traced lung adenocarcinoma, and mouse optic nerve
aging/intervention systems, spanning CosMx, Visium, Xenium, multiplexed
error-robust fluorescence in situ hybridization (MERFISH), single-cell RNA
sequencing (scRNA-seq), Slide-seq, Slide-tags, histology, and lineage-recording
data. Candidate claims are hardened through reproduction, independent scientist
review, and trajectory inspection.  Final answers are graded deterministically
over controlled vocabularies and symbols with companion rubrics capturing
progress through key analysis chokepoints. Across the SpatialBench-Long
benchmark, three model-harness pairs tie at 8/72 runs (11.1\%): Gemini 3.5
Flash / Pi terminal coding harness, GPT-5.5 / Pi, and GPT-5.5 / OpenAI Codex.
SpatialBench-Long tests whether agents can move beyond executing procedural
analysis to deriving accurate scientific conclusions from complex spatial
measurements.

\end{PreprintAbstract}

\vfill

\clearpage

\section*{Topline Benchmark Performance}

{\fontsize{9.1}{12.8}\selectfont
We ran the benchmark across frontier model families and agent harnesses. Pi denotes
the Pi terminal coding harness. Passing requires
exact recovery of the graded structured answer for an evaluation attempt. Present systems show low
but nonzero success rates: Gemini 3.5 Flash / Pi, GPT-5.5 / Pi, and GPT-5.5 /
OpenAI Codex each pass 8/72 runs. \par}

\begin{center}
  \includegraphics[width=0.94\textwidth]{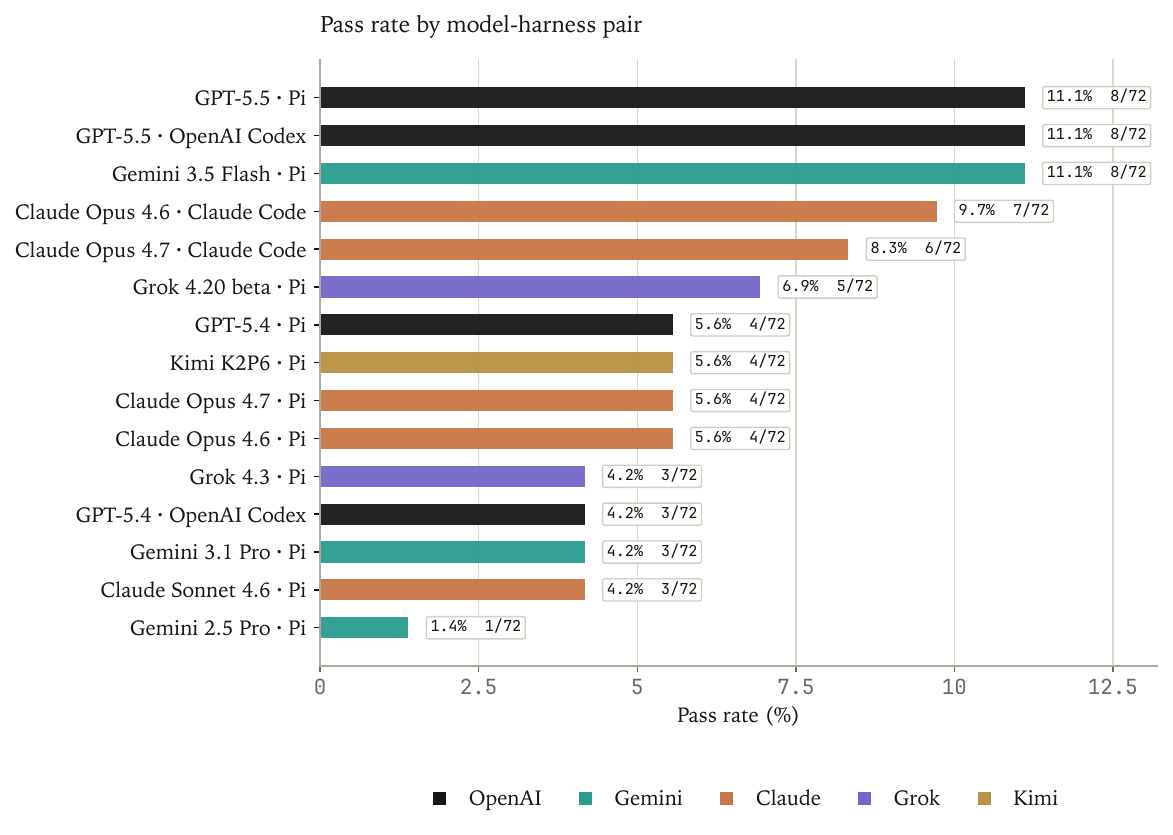}

  \vspace{0.18in}

  \begin{minipage}[t]{0.485\textwidth}
    \centering
    \includegraphics[width=\linewidth]{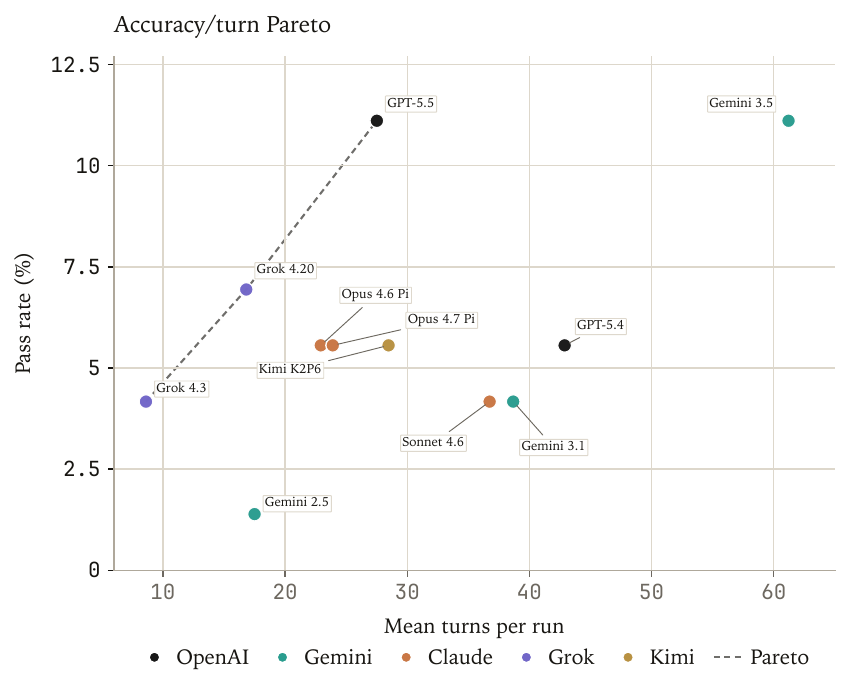}
  \end{minipage}
  \hfill
  \begin{minipage}[t]{0.485\textwidth}
    \centering
    \includegraphics[width=\linewidth]{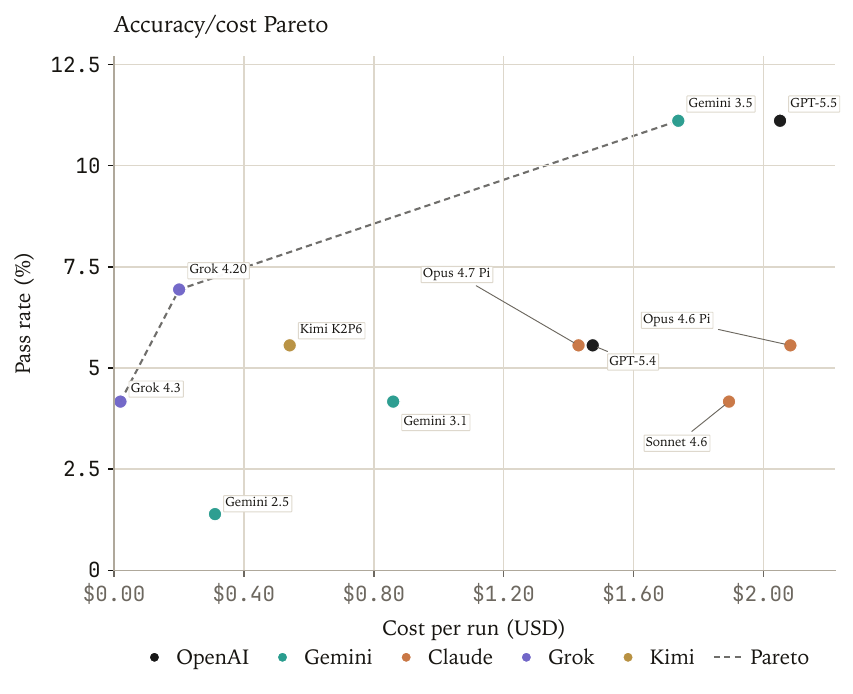}
  \end{minipage}

  \captionof{figure}{\textbf{Topline benchmark performance.} Top: pass rate by model-harness pair
  across the complete benchmark set. Bottom left: pass rate against mean agent turns.
  Bottom right: pass rate against mean cost.
  Confidence intervals and evaluation-level replicate statistics are reported in
  Table~\ref{tab:model-results}}
  \label{fig:topline-performance}
\end{center}

\StartBody

\section{Introduction}

Scientists use spatial biology data in open-ended research contexts to
construct new knowledge about living systems. Raw measurements do not directly
encode scientific conclusions. They must be processed through multi-step
workflows, integrated with other assays, interpreted against experimental
design, and contextualized with both prior literature and the original goal of
the study \cite{moses2022museum,williams2022intro,dries2021advances}.

AI agents are beginning to show utility in biological data analysis, but they
still struggle to contextualize analysis decisions within scientific goals, use
assay-specific knowledge accurately, and distinguish conclusions supported by
the provided data from plausible claims drawn from prior literature
\cite{workman2025spatialbench,workman2026scbench,qu2026biomnibench}.

Prior benchmarks focus on complementary aspects of scientific work: broad
biology reasoning benchmarks emphasize coverage, bioinformatics-agent
benchmarks emphasize executable workflows, and spatial-analysis benchmarks
emphasize deterministic grading of localized analysis steps
\cite{laurent2024labbench,mitchener2025bixbench,nair2026compbiobench,li2026genebench,anthropic2026biomysterybench,qu2026biomnibench,workman2025spatialbench,workman2026scbench}.
However, no current benchmark investigates whether an agent can undertake the
end-to-end work of recovering a specific scientific conclusion from raw spatial
measurements, calibrated context, and many possible analysis paths.

Raw spatial data rarely admits a single universal ground truth. The same
measurements and experimental context can support multiple valid scientific
conclusions, and unanticipated but true claims can be drawn from a dataset.
Published claims can also fail to reproduce cleanly under unbiased reanalysis
\cite{ioannidis2005false,prinz2011believe,begley2012standards,errington2021challenges}.

SpatialBench-Long evaluates whether agents can recover specified scientific
conclusions from raw spatial biology data and experimental context. Each task
provides a scientific question, raw or near-raw data, relevant assay context,
and a verifiable solution. The benchmark reports binary pass/fail grading on
final scientific conclusions and pairs verifiable scores with rubric-based
trajectory diagnostics that identify progress through key analysis chokepoints
\cite{workman2025spatialbench,workman2026scbench}. In the 24-evaluation
benchmark run set, the top three model-harness pairs each pass 8/72 runs, motivating
diagnostic analyses that distinguish total failure from partial scientific
progress.

\EndBody

\section{Benchmark Design}

{\fontsize{9.1}{12.8}\selectfont
Agents are evaluated on end scientific conclusions from 24 evaluations spanning
four study systems: primary pancreatic ductal adenocarcinoma (PDAC), engineered glioblastoma (GBM) organoids and in vivo
tumors, Cas9 lineage-traced lung adenocarcinoma, and mouse optic nerve aging
and intervention
\cite{lyubetskaya2026pdac,ishahak2025gbmorganoids,jones2024lineagetracing,groh2025microglia}.
Across the benchmark, evaluations draw on modalities including spatial
transcriptomics, histology, single-cell references, and lineage-recording data.
\par}

{\fontsize{9.1}{12.8}\selectfont
\begin{center}
\begin{tikzpicture}
  \node[
    fill=figurecream,
    draw=figureline,
    line width=0.45pt,
    rounded corners=8pt,
    inner xsep=0.20in,
    inner ysep=0.14in,
    text width=0.91\textwidth,
    align=left
  ] {%
    \begin{minipage}{0.91\textwidth}
      {\fontsize{8.5}{10}\selectfont\bfseries\color{paperaccent}Benchmark Construction Workflow}\par
      \vspace{0.08in}
      \centering
      \includegraphics[width=\linewidth]{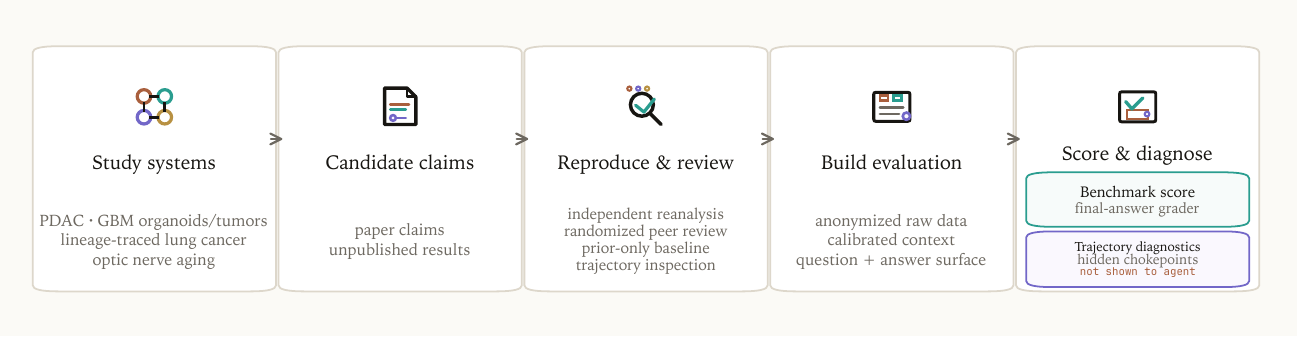}
      \captionof{figure}{\textbf{Benchmark construction workflow.}
      SpatialBench-Long tasks are built by selecting study systems,
      calibrating candidate claims through reproduction and review, packaging
      anonymized data and controlled answer surfaces, and pairing
      deterministic final-answer grading with hidden trajectory diagnostics.}
      \label{fig:claim-conditioned-eval}
    \end{minipage}
  };
\end{tikzpicture}
\end{center}
}

{\fontsize{9.1}{12.8}\selectfont
\begin{multicols}{2}

This organization reflects the structure of real spatial biology projects.
Solving these tasks requires cross-assay reasoning, experimental-design
awareness, and command of complex spatial biology workflows: tissue
segmentation, cell-neighborhood and niche analysis, spatial differential
expression, reference-based cell-type mapping, and histology–transcriptome
alignment
\cite{dries2021advances,singhal2024banksy,varrone2024cellcharter,qin2024spade,kleshchevnikov2022cell2location,biancalani2021tangram}.

Each evaluation includes experimental context, raw data, and a scientific
question. Task information is calibrated to approximate what a scientist would
know when beginning the analysis, balancing over-specification against
ambiguity. Task descriptions and raw-data labels are scrubbed for identifying
information about the original study. In many cases, the task tests an
unpublished result or is structured so that memorized literature alone is
unlikely to solve it
\cite{nair2026compbiobench,li2026genebench}.

Identifying durable ground truths is especially challenging in long-horizon
biological benchmarking. Naive paper-claim reproduction is brittle because the
same data may support multiple valid conclusions, and some published claims do
not reproduce cleanly under unbiased reanalysis
\cite{ioannidis2005false,prinz2011believe,begley2012standards,errington2021investigating}.
We therefore use paper claims as sources for candidate evaluations rather than
as automatic ground truth.
Candidate tasks are refined through independent reproduction, randomized expert
review, and inspection of trajectories from multiple model families. Many
candidate claims were excluded because they did not reproduce robustly from the
provided data.

Grading uses deterministic functions over structured final answers. Rather than
grading isolated numerical outputs from individual statistical operations,
SpatialBench-Long grades recovery of scientific conclusions expressed through
controlled biological vocabularies, ordered relationships or direction labels
\cite{workman2025spatialbench,workman2026scbench,nair2026compbiobench}.

Manual trajectory inspection is a first-class part of benchmark construction.
Model trajectories and randomized expert attempts are used to stress-test task
context, target answers, and grading assumptions.  Evaluation authors maintain
reproduction notes and rubrics describing known decision chokepoints. These
notes aid manual trajectory interpretation and provide a record for future
benchmark updates, especially as stronger models may solve tasks through
unanticipated but valid analysis paths that challenge current grading
assumptions \cite{qu2026biomnibench}.

\subsection{Verifiable grading paired with rubric diagnostics}

Benchmark scores use verifiable pass/fail grading on final outcomes. In
practice, failures to construct a verifiable grader usually indicated that a
candidate task lacked a reproducible target claim, that the provided context
was miscalibrated, or that the answer surface had not yet been constrained
enough for deterministic evaluation
\cite{workman2025spatialbench,workman2026scbench,nair2026compbiobench}.

However, final-answer grading provides sparse diagnostic signal for
long-horizon tasks. A model can fail the benchmark while solving many
subproblems correctly, and deterministic grading necessarily penalizes answers
outside the pre-specified target surface, including some valid claims not
anticipated by the benchmark authors
\cite{lightman2023verify,qu2026biomnibench}.

We considered verifiable grading of intermediate steps, but found that
specifying those steps risked leaking information about the intended solution
path or biasing the agent’s analysis trajectory. We therefore use rubric-based
trajectory judging as a companion diagnostic rather than as the benchmark
score. These rubric scores are prompt-sensitive and reflect the authors’
current model of the task, but they provide higher-resolution information about
partial progress and help identify cases where grading assumptions should be
revisited \cite{lightman2023verify,qu2026biomnibench}.

Evaluation authors define chokepoints after independent reproduction, peer
review, and inspection of trajectories from multiple model families.
Chokepoints are analysis decisions or biological constraints expected to remain
stable across plausible solution paths. Examples include choosing the correct
biological comparison, identifying the relevant assay or spatial compartment,
performing a necessary sanity check or avoiding a known trap.

Chokepoint rubrics are then used by large language model (LLM) judges to score model trajectories. We
report their correlation with verifiable pass/fail outcomes and their
consistency across replicates and model families, treating rubric scores as
diagnostic evidence rather than replacement benchmark scores
\cite{qu2026biomnibench,zheng2023llmjudge,wang2024fair,liu2023geval}.

\end{multicols}
}

\clearpage

\section[Motivating Example: Can an Agent Reconstruct A Primary Pro-Metastatic Cancer Niche]{Motivating Example: Can an Agent Reconstruct\\ A Primary Pro-Metastatic Cancer Niche}

{\fontsize{9.1}{12.8}\selectfont
\begin{center}
\begin{tikzpicture}
  \node[
    fill=figurecream,
    draw=figureline,
    line width=0.45pt,
    rounded corners=8pt,
    inner xsep=0.20in,
    inner ysep=0.14in,
    text width=0.91\textwidth,
    align=left
  ] {%
    \begin{minipage}{0.91\textwidth}
      {\fontsize{8.5}{10}\selectfont\bfseries\color{paperaccent}Lineage-tracing Chokepoint Schematic}\par
      \vspace{0.08in}
      \centering
      \includegraphics[width=\linewidth]{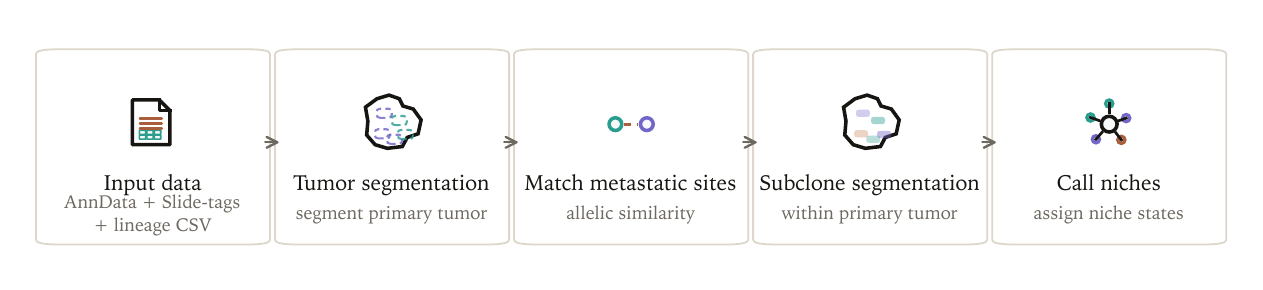}
      \captionof{figure}{\textbf{Primary pro-metastatic niche reconstruction.}
      The lineage-tracing task compresses a long spatial workflow into a
      verifiable structure. Agents must stage spatial and lineage inputs,
      segment candidate primary tumor regions, match metastatic references by
      allelic distance, define met-like subregions within the primary tumor,
      and call enriched or depleted niche programs while avoiding plausible
      generic metastasis signals.}
      \label{fig:lineage-chokepoint-flow}
    \end{minipage}
  };
\end{tikzpicture}
\end{center}

\begin{multicols}{2}

One evaluation begins with a compact biological question: which expression
programs characterize the primary-tumor regions most related to distant
metastases? The source study combines spatial transcriptomics with Cas9 lineage
tracing in a Kras;p53 lung adenocarcinoma model
\cite{jones2024lineagetracing,yang2022kptracer}. Cre induction both initiates
tumor formation and activates lineage recording, producing heritable
target-site edits that mark tumor subclones \cite{dupage2009conditional}.
Slide-seq and Slide-tags then measure spatial transcriptional state together
with lineage-target transcripts, making it possible to ask how clonal history
and local microenvironment vary across primary and metastatic lesions
\cite{rodriques2019slideseq,russell2024slidetags,jones2024lineagetracing}.

The agent is given anonymized spatial expression arrays of primary tumors and
metastatic lesions, lineage allele tables, distant metastatic references, and a
controlled vocabulary of candidate programs. It must reconstruct primary-tumor
niches most lineage-similar to distant metastases compared to non-metastatic
regions of the same primary tumor.

The evaluation authors identified a series of chokepoints. The agent must first
stage the spatial and lineage data together instead of relying only on gene
expression data; segment candidate primary tumor regions without falling
outside shape and count uncharacteristic of lung cancer; use allelic distance
to connect metastatic references back to primary regions instead of relying
expression similarity; define a spatially coherent metastatic subregion within
the primary tumor; and finally call niche programs consistent across multiple
layers of the primary tumor and separating the primary pro-metastatic niche
from distant-site remodeling or generic metastasis signatures.

The deterministic grader scores only the final primary tumor niches with the
correct directions. The accepted enriched programs include
epithelial-to-mesenchymal transition (EMT), hypoxia, immunosuppressive macrophage, scavenger macrophage, and fibrotic
programs, while alveolar differentiation and gastric/endoderm programs are
depleted.  Nearby alternatives such as collagen deposition, myogenesis,
blood-vessel programs, and generic metastasis signatures are biologically
plausible but rejected or ambiguous for this specific primary-niche claim after
analyzing multiple analysis paths.

This example shows why rubric scores are useful even when the benchmark score
is binary. A model can trip up in the final stages of the evaluation, such as
only considering a single primary tumor layer and picking a distractor niche
with an incorrect local direction while making considerable correct progress.

\end{multicols}
}

\clearpage

\section{Evaluation Inventory}

{\fontsize{9.1}{12.8}\selectfont
SpatialBench-Long is organized around studies instead of independent datasets.
Each study contributes a cluster of evaluations that asks different scientific
questions of the same experimental context while varying the relevant assays
and tasks.

Across the four studies, the benchmark currently contains 24 long-horizon evaluations. Primary
PDAC contributes the largest cluster, with tasks over CosMx, Visium data from formalin-fixed
paraffin-embedded (FFPE) tissue, paired CosMx/Visium evidence, and histology. Engineered glioblastoma
(GBM) organoid tasks combine Xenium, matched single-cell RNA sequencing (scRNA-seq),
and patient GBM single-cell references. Lung adenocarcinoma tasks pair Slide-seq and Slide-tags
spatial measurements with Cas9 lineage recording. Optic nerve tasks use multiplexed
error-robust fluorescence in situ hybridization (MERFISH) single-cell spatial transcriptomics
in aging and intervention settings. H\&E denotes hematoxylin and eosin staining; CAF, TME,
and TLS denote cancer-associated fibroblast, tumor microenvironment, and tertiary lymphoid
structure, respectively.\par}

\begin{center}
\fontsize{7.2}{8.7}\selectfont
\setlength{\tabcolsep}{3.2pt}
\begin{tabular}{@{}>{\raggedright\arraybackslash}p{0.24\linewidth}
                >{\centering\arraybackslash}p{0.055\linewidth}
                >{\raggedright\arraybackslash}p{0.27\linewidth}
                >{\raggedright\arraybackslash}p{0.38\linewidth}@{}}
\toprule
Study/system & Evals & Assays / data types & Main task themes \\
\midrule
Primary PDAC &
11 &
CosMx, Visium FFPE, paired CosMx+Visium, H\&E/trichrome &
Classical-basal axis, CAF/immune/TME shifts, hypoxia, TLS, perineural niches, collagen and histology integration \\
\addlinespace
Glioblastoma organoids / in vivo tumors &
4 &
Xenium, matched scRNA-seq, patient GBM single-cell RNA reference &
Organoid niche disruption, proximity changes, in vivo tumor architecture, malignant cell-state composition \\
\addlinespace
Lung adenocarcinoma lineage tracing &
5 &
Slide-seq, Slide-tags, Cas9 lineage recording &
Metastasis-origin matching, metastatic-site programs, tumor fitness, pro-metastatic primary niches \\
\addlinespace
Mouse optic nerve aging/intervention &
4 &
MERFISH single-cell spatial transcriptomics &
Aging niche shifts, chemokine source attribution, spatial colocalization, condition-anonymized disruption \\
\bottomrule
\end{tabular}
\captionof{table}{\textbf{Evaluation inventory.} SpatialBench-Long groups
    evaluations by study while varying the claim, workflow, data types, and
    task objectives.}
\label{tab:evaluation-inventory}
\end{center}

{\fontsize{8.3}{11.4}\selectfont
Across these evaluations, assay coverage includes CosMx-only tasks, Visium and
Visium-plus-histology tasks, mixed CosMx/Visium tasks, Xenium and scRNA-seq
reference tasks, MERFISH tasks, and Slide-seq/Slide-tags/Cas9-lineage tasks.
The task set emphasizes spatial niche and architecture reasoning,
cross-platform reconciliation, tumor-state and microenvironment interpretation.
The complete benchmark reports 15 model-harness pairs and 72 runs per pair
across the 24 evaluations, for 1,080 trajectories.
\par}

\StartBody

\section{Results}

\subsection{Verifiable grading stratifies frontier models with low pass rates}

Across 15 model-harness pairs and 1,080 trajectories, Gemini 3.5 Flash / Pi,
GPT-5.5 / Pi, and GPT-5.5 / OpenAI Codex each passed 8/72 attempts
(11.11\%; Wilson 95\% confidence interval (CI), 5.74--20.42), with Claude Opus 4.6 / Claude Code
close behind at 7/72 attempts (9.72\%). Most remaining model-harness pairs
passed 3--5/72 attempts, while Gemini 2.5 Pro / Pi passed 1/72. Success was
therefore low across all systems and not cleanly separated among the leading
models. Evaluation-level replicate statistics help distinguish occasional
success from clearer task recovery: even among the leading systems, passing at
least one replicate was observed for only 4--5 of 24 evaluations, and
majority-replicate passing for only 2 of 24
(Figure~\ref{fig:main-results-summary}; Table~\ref{tab:model-results}).

\EndBody

\begin{center}
  \includegraphics[width=0.94\textwidth]{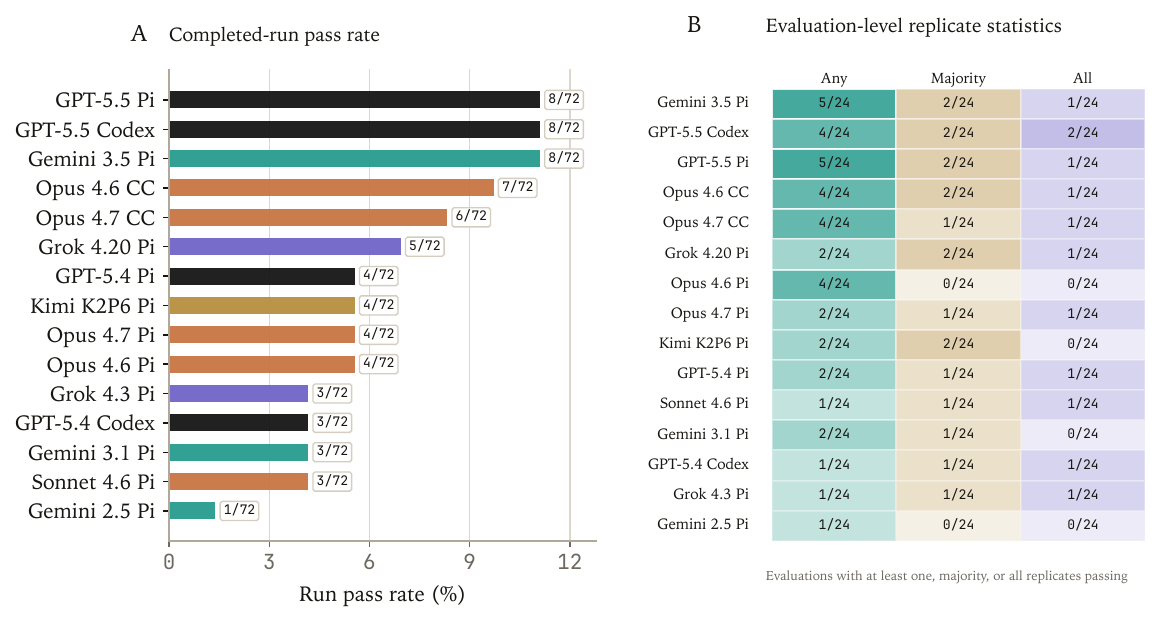}

  \captionof{figure}{\textbf{Sparse endpoint success across model-harness pairs.}
  Deterministic final-answer performance across 24 SpatialBench-Long evaluations.
  The left panel reports passing trajectories out of 72 attempts per model-harness
  pair. The right panel reports evaluation-level replicate statistics: the number
  of evaluations with at least one passing replicate, a majority of passing
  replicates, or all replicates passing. Wilson confidence intervals are omitted
  from the figure and reported in Table~\ref{tab:model-results}.}
  \label{fig:main-results-summary}
\end{center}

\vspace{0.08in}

\begin{center}
{\fontsize{5.2}{6.4}\selectfont
\begin{tabular}{@{}lrrrrrrrr@{}}
\toprule
Model / harness & Pass runs & Pass \% & Wilson 95\% CI & Any & Majority & All & Cost & Turns \\
\midrule
Gemini 3.5 Flash / Pi & 8/72 & 11.11 & 5.74--20.42 & 5/24 & 2/24 & 1/24 & \$1.7378 & 61.21 \\
GPT-5.5 / OpenAI Codex & 8/72 & 11.11 & 5.74--20.42 & 4/24 & 2/24 & 2/24 & -- & -- \\
GPT-5.5 / Pi & 8/72 & 11.11 & 5.74--20.42 & 5/24 & 2/24 & 1/24 & \$2.0512 & 27.51 \\
Claude Opus 4.6 / Claude Code & 7/72 & 9.72 & 4.79--18.74 & 4/24 & 2/24 & 1/24 & -- & -- \\
Claude Opus 4.7 / Claude Code & 6/72 & 8.33 & 3.88--17.01 & 4/24 & 1/24 & 1/24 & -- & -- \\
Grok 4.20 beta / Pi & 5/72 & 6.94 & 3.00--15.25 & 2/24 & 2/24 & 1/24 & \$0.2004 & 16.82 \\
Claude Opus 4.6 / Pi & 4/72 & 5.56 & 2.18--13.43 & 4/24 & 0/24 & 0/24 & \$2.0830 & 22.92 \\
Claude Opus 4.7 / Pi & 4/72 & 5.56 & 2.18--13.43 & 2/24 & 1/24 & 1/24 & \$1.4305 & 23.90 \\
Kimi K2P6 / Pi & 4/72 & 5.56 & 2.18--13.43 & 2/24 & 2/24 & 0/24 & \$0.5409 & 28.47 \\
GPT-5.4 / Pi & 4/72 & 5.56 & 2.18--13.43 & 2/24 & 1/24 & 1/24 & \$1.4742 & 42.88 \\
Claude Sonnet 4.6 / Pi & 3/72 & 4.17 & 1.43--11.55 & 1/24 & 1/24 & 1/24 & \$1.8940 & 36.74 \\
Gemini 3.1 Pro / Pi & 3/72 & 4.17 & 1.43--11.55 & 2/24 & 1/24 & 0/24 & \$0.8596 & 38.67 \\
GPT-5.4 / OpenAI Codex & 3/72 & 4.17 & 1.43--11.55 & 1/24 & 1/24 & 1/24 & -- & -- \\
Grok 4.3 / Pi & 3/72 & 4.17 & 1.43--11.55 & 1/24 & 1/24 & 1/24 & \$0.0195 & 8.61 \\
Gemini 2.5 Pro / Pi & 1/72 & 1.39 & 0.25--7.46 & 1/24 & 0/24 & 0/24 & \$0.3108 & 17.50 \\
\bottomrule
\end{tabular}
}

\captionof{table}{\textbf{Deterministic final-answer performance.}
Model-harness results across the 24-evaluation SpatialBench-Long benchmark.
Pass runs and pass percentage report endpoint success over 72 attempts per
model-harness pair. Wilson intervals summarize run-level binomial uncertainty
and do not account for evaluation-level clustering; evaluation-level replicate
statistics are reported separately in the Any, Majority, and All columns. Cost
and turn-count summaries are shown where comparable metadata are available;
code-harness result JSONs lacking these fields are shown with dashes.}
\label{tab:model-results}
\end{center}

\clearpage

{\fontsize{9.1}{12.8}\selectfont
\begin{multicols}{2}

\subsection{Rubric Judges Are Reproducible and Associated with Endpoint Success}

Rubric judges are used as companion diagnostics to measure partial progress and
interpret model behavior against sparse final-answer binary grading. We scored
792 trajectories with four judges, yielding 3,145 valid scores out of 3,168
expected scores (99.27\%). Of these trajectories, 770 had valid scores from all
four judges and 791 had valid scores from at least three.

Rubric scores were associated but not perfectly correlated with endpoint
success. Among trajectories with four valid judge scores, endpoint-passing runs
had higher mean rubric scores than endpoint-failing runs (72.2\%, bootstrap
95\% CI 62.1--80.9, vs. 45.8\%, 44.1--47.5). The four-judge rubric mean showed
a modest association with deterministic pass/fail status (Pearson $r=0.24$,
bootstrap 95\% CI 0.16--0.33; Spearman $\rho=0.24$; receiver operating characteristic area under the curve (ROC AUC) $=0.79$, bootstrap 95\%
CI 0.70--0.88; Figure~\ref{fig:judge-validation}).

Judge-to-judge reproducibility was high. Mean pairwise judge correlation was
0.93, the minimum pairwise correlation was 0.91, and the mean absolute pairwise
score difference was 6.2 percentage points. Same-trajectory cross-judge
variability was smaller than variation across independent attempts on the same
evaluation: judge standard deviation (SD) averaged 5.3 points per trajectory, while replicate SD
averaged 8.7 points across eval-by-source-model groups with at least two
trajectories (Figure~\ref{fig:judge-validation}).

These results support rubric scores as reproducible diagnostic annotations: they
are informative but imperfect predictors of curated endpoint success, not
substitutes for verifiable final-answer grading.

\end{multicols}
}

\vspace{0.04in}

\begin{center}
  \includegraphics[width=0.90\textwidth]{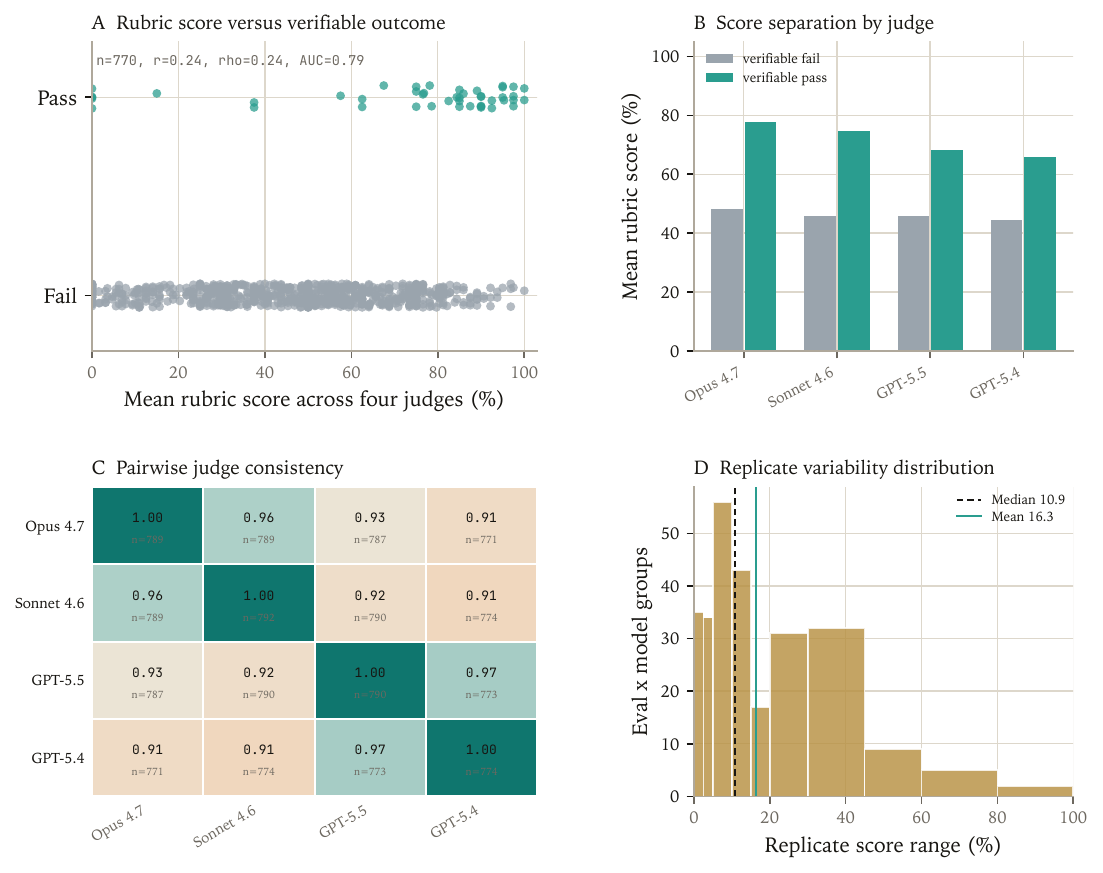}

\captionof{figure}{\textbf{Rubric diagnostics are reproducible and endpoint-associated.}
Rubric-judge analysis across the 792-trajectory judge matrix. Panel A compares
mean four-judge rubric score with deterministic pass/fail outcomes for
trajectories with all four valid judge scores. Panel B shows the same
pass/fail separation by judge. Panel C reports pairwise judge-score correlations
on matched trajectories. Panel D shows rubric-score variation across independent
attempts from the same source model and evaluation. Rubric scores are used as
diagnostic trajectory annotations, not as benchmark pass/fail scores.}
\label{fig:judge-validation}
\end{center}

\StartBody

\subsection{Rubric Scores Are Useful but Do Not Replace Verifiable Grading}

Strict endpoint grading makes reward sparse. In the 792-trajectory judge matrix,
only 47 trajectories passed the deterministic grader (5.9\%).  Replicate pass
rates were more informative but still zero for most trajectories, with nonzero
scores for 275/714 trajectories (38.5\%).  By contrast, mean rubric score was
nonzero for 725/770 trajectories (94.2\%), indicating that rubric judging
supplies much denser trajectory-level signal than endpoint grading
(Figure~\ref{fig:rubric-reward-reliability}).

We were interested in exploring the utility of rubric grading for intermediate
reward and asked if denser signal was partially aligned with endpoint quality.
Mean four-judge rubric score correlated with replicate pass rate score at
Pearson $r=0.361$ (eval-cluster bootstrap 95\% CI 0.074--0.573; $n=692$) and
Spearman $\rho=0.338$ (0.085--0.552). 

To check that this association was not driven only by differences in evaluation
difficulty, we repeated the analysis after centering scores within each
evaluation. The correlation remained similar ($r=0.340$, 0.149--0.510),
indicating that higher-scored trajectories were more likely to succeed even
among attempts on the same evaluation.

The relationship between endpoint grading and rubrics was uneven. Verifiable
scores were enriched in the highest rubric decile, where the pass rate reached
34.2\%, but pass rates were noisy and nonmonotonic across the middle deciles.
Thus, very high rubric scores identify trajectories enriched for endpoint
success, while small or moderate rubric-score differences should not be
interpreted as calibrated differences in scientific quality.

Rubric scores also varied systematically by source model. After controlling for
evaluation identity, normalized verifiable score, and pass/fail status, mean
residual rubric scores still differed substantially across source models and
model families. Source-model residuals ranged from -26.8 to +11.6 percentage
points, while source-family residuals ranged from -20.9 to +11.2 points. Raw
rubric scores therefore appear to reflect not only endpoint success, but also
trajectory style, model family, and possible judge-style effects
(Figure~\ref{fig:rubric-reward-reliability}).

Together, these analyses support rubric scores as promising auxiliary tools,
not substitutes for verifiable endpoint grading. High rubric scores enrich for
endpoint success, but moderate score differences are not calibrated measures of
scientific correctness. Stronger validation requires expert trajectory audits
to decide whether deviations are valid alternate analyses, rubric over-credit,
or ordinary model errors.

\EndBody

\begin{center}
  \includegraphics[width=0.94\textwidth]{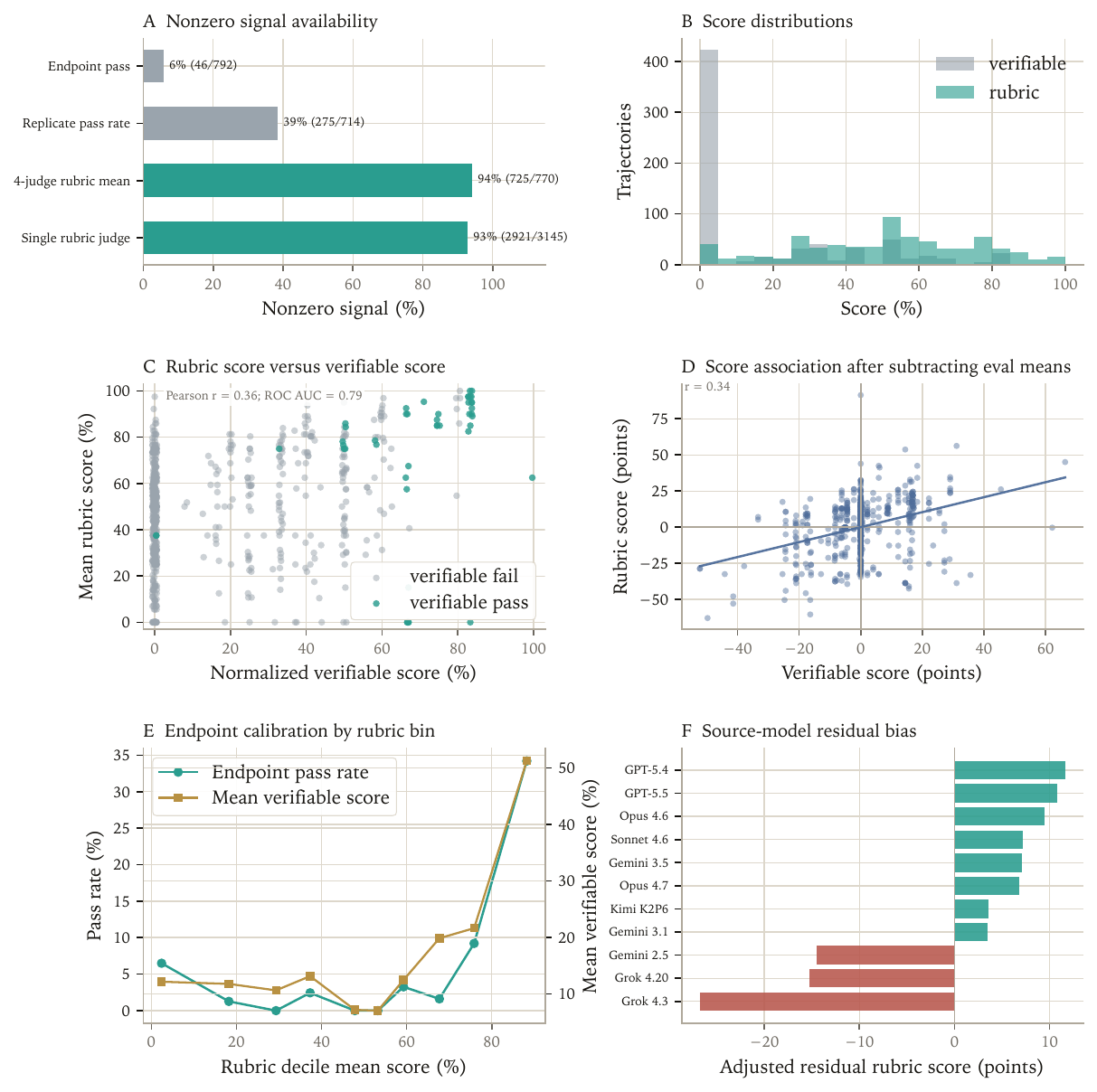}

\captionof{figure}{\textbf{Rubric scores provide dense but imperfect auxiliary signal.}
Panel A compares nonzero rates for deterministic endpoint passes,
replicate-pass rate, mean rubric score, and single-judge rubric score.
Panel B compares the corresponding score distributions. Panel C shows the
relationship between mean rubric score and replicate pass rate for the
corresponding evaluation-by-source-model group. Panel D shows the same
relationship after subtracting each evaluation's mean from both scores. Panel E
reports endpoint pass rate by rubric-score decile. Panel F reports source-model
residual rubric variation after controlling for evaluation identity and
normalized verifiable score, and pass/fail status.
Rubric scores are denser than endpoint grading, but are not calibrated
substitutes for verifiable pass/fail outcomes.}
\label{fig:rubric-reward-reliability}
\end{center}

\StartBody

\subsection{Rubric Scores Vary by Source Model and Trajectory Style}

We next stratified rubric scores by the source agent model whose trajectory was
being judged to see if rubric patterns are consistent across judges.  Mean
four-judge rubric scores were highest for OpenAI and Anthropic trajectories.
GPT-5.4 and GPT-5.5 both averaged about 63\%, while Claude Opus 4.6, Claude
Opus 4.7, and Claude Sonnet 4.6 averaged about 60\%, 56\%, and 56\%,
respectively. Gemini 3.5 Flash had one of the strongest verifiable profiles but
a lower mean rubric score (46.5\%), suggesting brittleness in either our
rubrics or ground truths. Kimi, Grok 4.20, Gemini 2.5 Pro, and Grok 4.3 had
lower mean rubric scores, ranging from 38.3\% to 23.4\%
(Figure~\ref{fig:judge-agent-model}).

These source-model patterns were consistent across judges. All four judges
preserved the same broad ordering across source models, while differing
modestly in absolute score level. Mean four-judge rubric score also correlated
with replicate-level endpoint signal (Pearson $r=0.36$; Spearman $\rho=0.34$;
$n=692$), but within-model correlations were descriptive rather than
inferential because endpoint passes were sparse and each model family produced
a different mix of failure modes.

\EndBody

\begin{center}
  \includegraphics[width=0.94\textwidth]{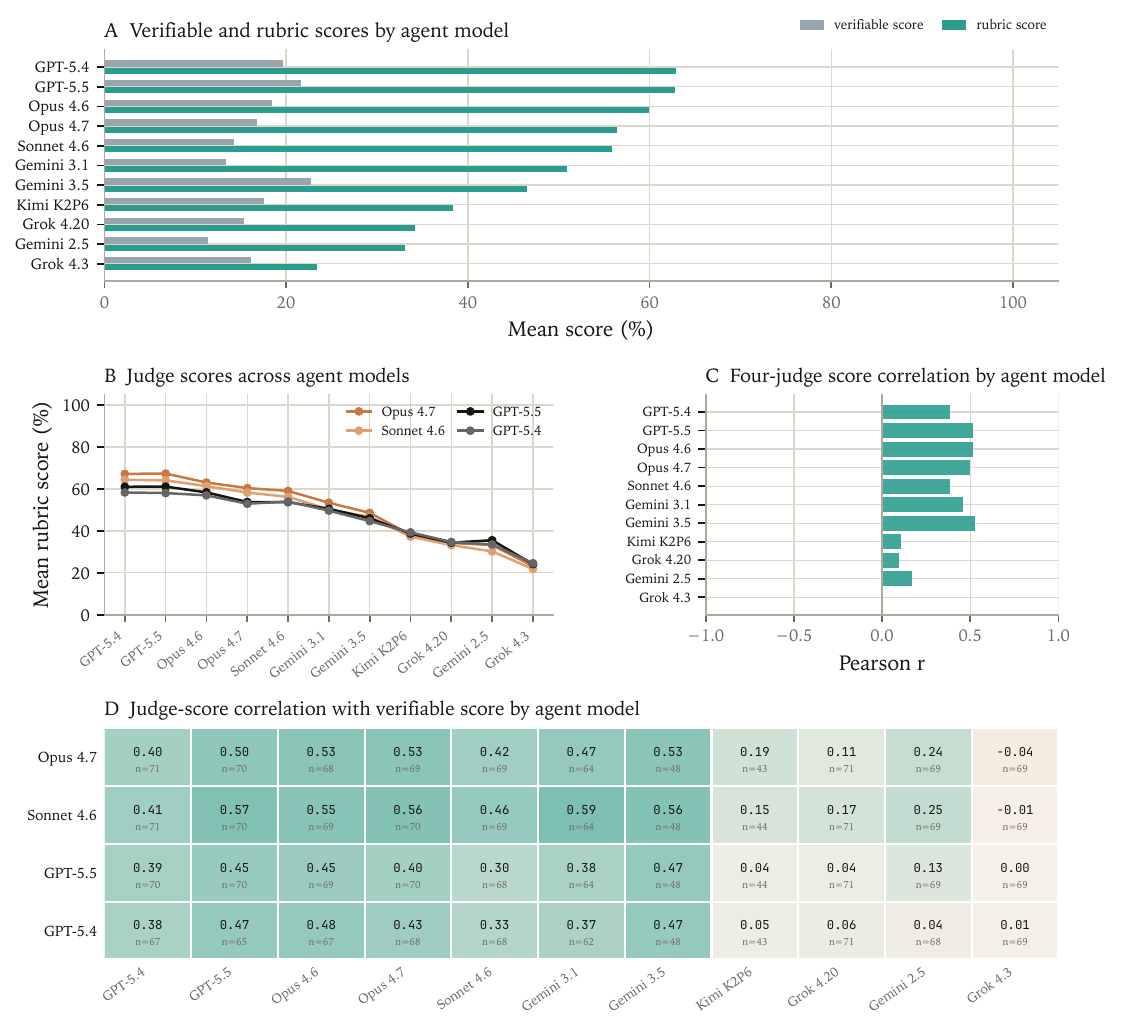}

\captionof{figure}{\textbf{Rubric scores vary by source agent model.} Panel A
    compares replicate-level endpoint signal with mean four-judge rubric score
    across source agent models. Panel B shows judge-specific rubric scores by
    source model. Panel C reports within-model correlations between mean
    four-judge rubric score and replicate-level endpoint signal. Panel D shows
the judge-by-agent-model correlation matrix.}
\label{fig:judge-agent-model}
\end{center}

\StartBody

\subsection{Model Behavior Descriptions}

Manual trajectory review identified seven recurring behavior groups. Two groups
with high rubric scores involved failures after doing most of the work and
collecting the requisite context. In final-decision failures, agents selected
the wrong final target. In partial-credit answers, agents returned some
accepted components but lost credit through omissions, aliases,
controlled-vocabulary mismatch, or distractor selection.

The remaining modes reflected deeper analysis failures. Twelve trajectories
missed important metadata or sanity-check context, most clearly in adult/aged
optic-nerve tasks where agents consistently missed intentionally swapped
metadata labels. Ten used the wrong grouping variable, especially in lineage
tasks where agents used expression intensity, whole-allele overlap, or generic
primary-versus-metastasis contrasts instead of lineage-aware comparisons. Six
used an inappropriate spatial method by clustering expression rather than
neighborhoods, pooling coordinates across sections, or annotating phantom cell
types. Six relied on prior biology or vocabulary hints instead of analyzing the
data. Three failed through local computation or data-wrangling issues.

Failure modes also differed by source model family. OpenAI and Anthropic
trajectories had the highest mean rubric score and most often failed at the
final decision after substantial analysis. Gemini failures often involved the
wrong grouping variable, threshold rules, or data wrangling. xAI and Kimi
failures skewed toward more shallow analysis.

\EndBody

\begin{center}
  \includegraphics[width=0.94\textwidth]{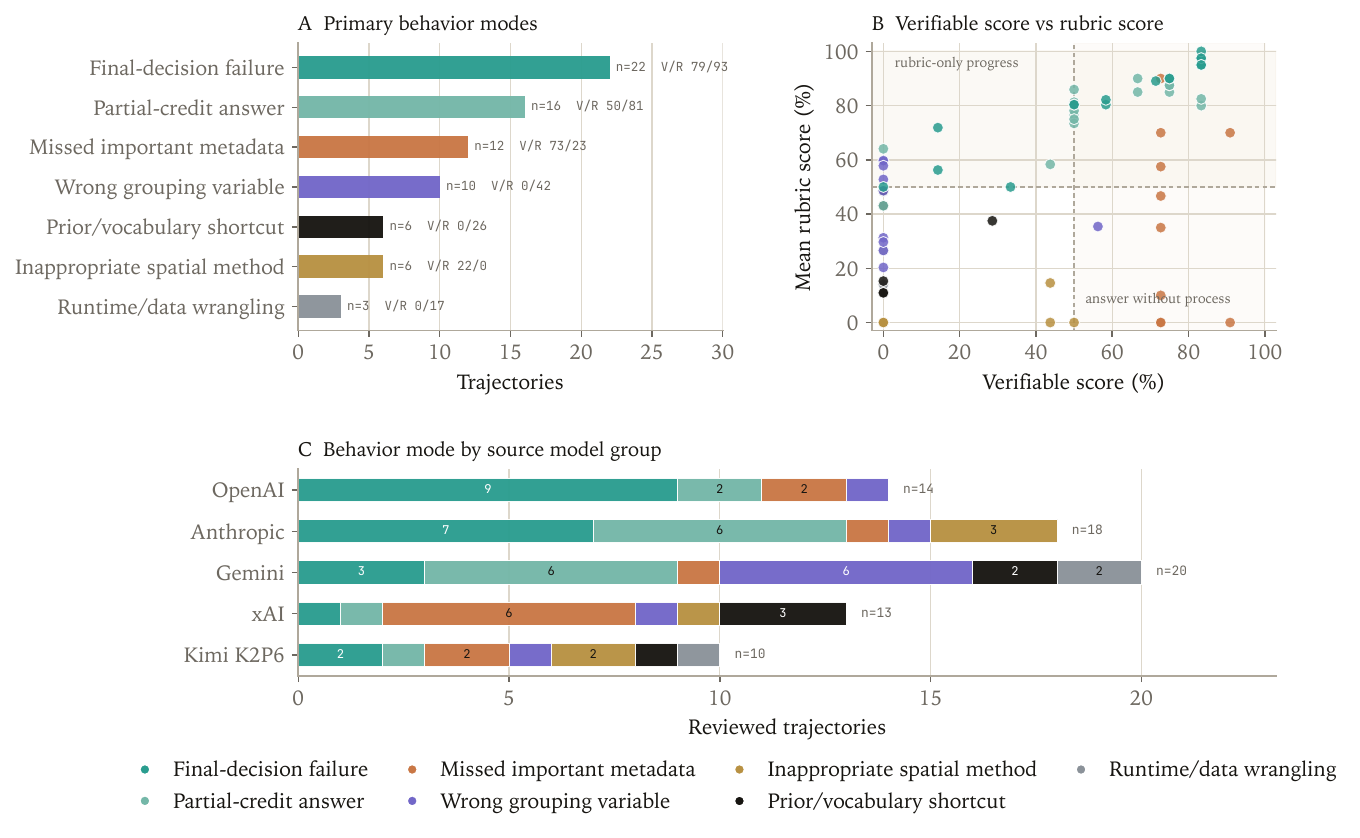}

\captionof{figure}{\textbf{Trajectory behavior modes.} Panel A summarizes the primary behavior
mode assigned to each of the 75 reviewed calibration trajectories. Inline labels report the count
and median verifiable/rubric score percentages for each group. Panel B compares normalized
verifiable score with mean rubric score across the same reviewed trajectories, showing how rubric
diagnostics separate shallow shortcuts, missed metadata, partial-credit answers, and final-decision
failures that can look similar under final-answer grading alone. Panel C stratifies the same
behavior modes by source model group.}
\label{fig:model-behavior}
\end{center}

\StartBody

\section{Discussion}
SpatialBench-Long evaluates whether AI agents can recover specific scientific
claims from raw spatial biology measurements and calibrated experimental
context. The benchmark preserves the open-ended structure of real analysis:
agents are not given prescribed methods, but must choose comparisons, process
assay-specific data, integrate spatial and molecular evidence, and return a
constrained scientific conclusion. This design makes evaluation more difficult
than procedural benchmarks, but exposes whether agents can coordinate many
local analysis decisions into a correct biological claim.

Current frontier agents show low but nonzero capability on this benchmark. The
leading model-harness pairs each passed at least one replicate on only 4--5 of
24 evaluations, and achieved majority-replicate success on only 2 of 24
evaluations, despite each passing 8 of 72 total attempts. Success was therefore
sparse at both the run level and the evaluation level. Rubric diagnostics and
trajectory review suggest that compounding local analysis errors, rather than
absence of general biological knowledge, prevent reliable long-horizon
scientific reasoning.

Before models can reliably reason about disease mechanisms, drug response, or
other deep results in biology, they must become procedurally competent at
concrete molecular measurements. In spatial biology, that means understanding
the properties and constraints of individual kits, machines, panels, tissue
types, coordinate systems, segmentation outputs, normalization regimes, and
experimental designs. Just as coding agents had to become reliable at local
software tasks before attempting larger systems, biology agents will likely
progress from assay-specific analysis competence to emergent biological
reasoning, and only later to synthesis across modalities, translational
context, and realistic ambiguity. Benchmarks such as SpatialBench-Long are
intended to make that progression measurable.

\section{Methods}

\subsection{Benchmark assembly}

Benchmark construction is described in the Benchmark Design section. Briefly,
we assembled 24 evaluations from four study systems spanning primary PDAC,
engineered glioblastoma organoids and in vivo tumors, Cas9 lineage-traced lung
adenocarcinoma, and mouse optic nerve aging/intervention. Candidate evaluations
were retained when the target claim could be reproduced from the provided data,
expressed through a constrained answer surface, and graded deterministically.
Tasks were excluded when the claim was not reproducible, the context was
insufficient or overdetermined, or the answer surface could not be made stable.
Each retained evaluation includes a prompt, raw or near-raw data, controlled
vocabulary, output schema, hidden grader, and task-specific rubric. Dataset
labels and study context were anonymized or scrubbed where needed to reduce
memorization and literature-lookup shortcuts.

\subsection{Agent runs}

We evaluated 15 model-harness pairs across Pi (the Pi terminal coding harness), OpenAI Codex, and Claude Code
harnesses. Each pair was run on all 24 evaluations with three independent
replicates, yielding 72 runs per pair and 1,080 final trajectories. Runs used
the packaged evaluation inputs, hidden graders, a shared execution environment,
tool access for file inspection and analysis, and fixed resource/time limits.
For each run we recorded the final answer, grader output, trajectory, logs,
duration, and available cost/turn metadata. Failed, timed-out, malformed, or
incomplete runs were retained in denominators and counted as nonpassing unless
the deterministic grader emitted a passing result. Cost and turn analyses are
restricted to the Pi harness, where those fields were consistently logged.

\subsection{Endpoint grading}

The primary benchmark score is deterministic endpoint grading of the final
structured answer. Graders parse the submitted output, compare fields against
task-specific schemas and controlled vocabularies, and compute run-level
pass/fail plus available component scores. Malformed, missing, or unparsable
answers are counted as failures. We aggregate endpoint results as run-level
pass rates and as evaluation-level replicate passing counts: the number of
evaluations for which a model-harness pair passes any replicate, a majority of
replicates, or all replicates.

\subsection{Rubric judging}

For diagnostic analysis, each evaluation has a chokepoint rubric describing
expected scientific progress through key analysis decisions and known traps.
Rubric scores are not benchmark scores. We scored the Pi-harness trajectory
matrix with four judge models: Opus 4.7, Sonnet 4.6, GPT-5.5, and GPT-5.4,
yielding 3,168 expected judge scores over 792 trajectories. Judges saw the
trajectory, task context, and rubric, and returned structured criterion scores.
Scores were normalized by rubric maximum and averaged across valid judges.
Invalid or missing judge outputs were excluded from judge-specific analyses;
matched-judge analyses use trajectories with the required valid scores.

\subsection{Statistical analysis}

Run-level pass rates use Wilson binomial confidence intervals. Rubric analyses
use bootstrap confidence intervals, clustered by evaluation when trajectories
from the same task are compared. Associations between rubric and endpoint
scores are reported with Pearson and Spearman correlations; pass/fail
discrimination is reported with ROC AUC. For
within-evaluation analyses, rubric and verifiable scores were centered by
evaluation to ask whether rubric scores rank trajectories within the same task.
Residual rubric-bias analyses control for evaluation identity, normalized
verifiable score, and pass/fail status.

\subsection{Manual trajectory review}

We manually reviewed 75 trajectories selected to cover passes, ordinary
failures, high-rubric failures, low-rubric passes, high judge-disagreement
cases, and high replicate-variation groups. Reviewers assigned qualitative
behavior modes: final-decision failure, partial-credit answer, missed important
metadata, wrong grouping variable, inappropriate spatial method, prior/vocabulary
shortcut, or runtime/data-wrangling breakdown. Reviewers also flagged possible
grader, rubric, or ground-truth issues separately from ordinary model failures.

\section*{Data availability}

Results files and a subset of evaluations and trajectories are available at
\url{https://github.com/latchbio/spatialbench-long}.

\EndBody

\end{document}